%% file: emnlp2018.tex
\documentclass[11pt,a4paper]{article}
\usepackage[hyperref]{emnlp2018}
\usepackage{times}
\usepackage{latexsym}
\usepackage{comment}
\usepackage{graphicx}
\usepackage{tabularx,ragged2e}
\usepackage{booktabs}
\usepackage{amsmath,amssymb,amsfonts} %
\usepackage{calligra}
\usepackage{bbm}
\usepackage{color}
\usepackage{url}
\usepackage{lipsum}

\newcommand{\lisa}[1]{\textcolor{red}{Lisa: #1}}

\newcommand{\anja}[1]{\textcolor{green}{Anja: #1}}

\newcommand\blfootnote[1]{%
  \begingroup
  \renewcommand\thefootnote{}\footnote{#1}%
  \addtocounter{footnote}{-1}%
  \endgroup
}

\newcommand{\metric}{CHAIR}
\newcommand{\metricfullname}{Caption Hallucination Assessment with Image Relevance}
\aclfinalcopy %
\title{Object Hallucination in Image Captioning}

\author{\textbf{Anna Rohrbach}$^{*1}$,
\textbf{Lisa Anne Hendricks}$^{*1}$,\\
\textbf{Kaylee Burns}$^{1}$ ,
\textbf{Trevor Darrell}$^1$,
\textbf{Kate Saenko}$^2$\\
$^1$ UC Berkeley, $^2$ Boston University \\
}

\date{}
\hypersetup{draft}
\begin{document}
\maketitle
\blfootnote{* Denotes equal contribution.}
\begin{abstract} 
Despite continuously improving performance, contemporary image captioning models are prone to ``hallucinating'' objects that are not actually in a scene. One problem is that standard metrics only measure similarity to ground truth captions and may not fully capture image relevance. In this work, we propose a new image relevance metric to evaluate current models with veridical visual labels and assess their rate of object hallucination. We analyze how captioning model architectures and learning objectives contribute to object hallucination, explore when hallucination is likely due to image misclassification or language priors, and assess how well current sentence metrics capture object hallucination. We investigate these questions on the standard image captioning benchmark, MSCOCO, using a diverse set of models. 
Our analysis yields several interesting findings, including that models which score best on standard sentence metrics do not always have lower hallucination and that models which hallucinate more tend to make errors driven by language priors.
\end{abstract}

\input{introduction}

\input{method}

\input{experiments}

\input{discussion}

\bibliography{emnlp2018}
\bibliographystyle{acl_natbib_nourl}

\end{document}

%% file: introduction.tex
\section{Introduction}
\label{sec:intro}

\begin{figure}[t]
\centering
\includegraphics[width=\linewidth]{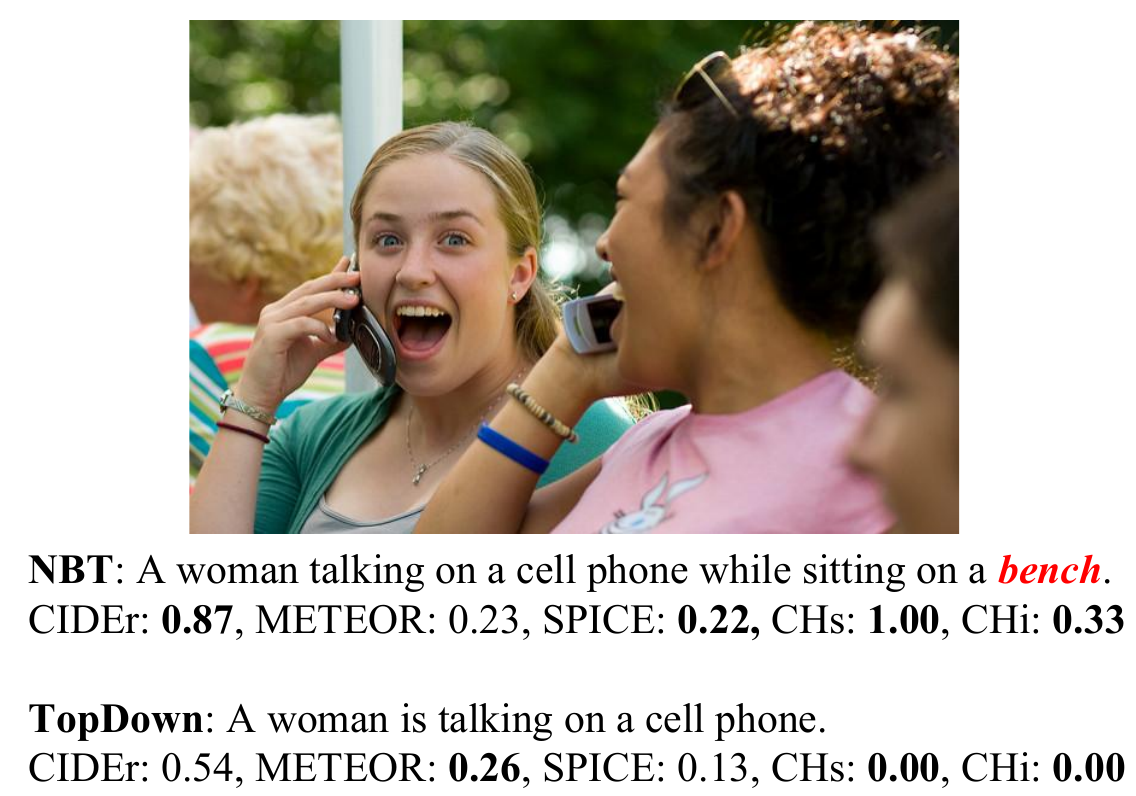}
\caption{\label{fig:teaser} Image captioning models often ``hallucinate'' objects that may appear in a given  context, like e.g. a \emph{bench} here. Moreover, the sentence metrics do not always appropriately penalize such hallucination.  Our proposed metrics (\metric{}s and \metric{}i) reflect hallucination.  For \metric{} \textit{lower is better}.
}
\end{figure}

Image captioning performance has dramatically improved over the past decade. Despite such impressive results, it is unclear to what extent captioning models actually rely on image content: as we show, existing metrics fall short of fully capturing the captions' relevance to the image. In Figure~\ref{fig:teaser} we show an example where a competitive captioning 
model, Neural Baby Talk (NBT)~\cite{lu2018neural}, incorrectly generates the object ``bench.'' 
We refer to this issue as object \emph{hallucination}.

While missing salient objects is also a failure mode, captions are summaries and thus generally not expected to  describe all objects in the scene. On the other hand, describing objects that are \emph{not present} in the image has been shown to be less preferable to humans.
For example, the LSMDC challenge~\cite{lsmdc-2017}
documents that correctness is more important to human judges than specificity. 
In another study, ~\cite{macleod} analyzed how visually impaired people react to automatic image captions. They found that people vary in their preference of either coverage or correctness. For many visually impaired who value correctness over coverage, hallucination is an obvious concern.
Besides being poorly received by humans, object hallucination reveals an internal issue of a captioning model, such as not learning a very good representation of the visual scene or overfitting to its loss function. 

In this paper we assess the phenomenon of object hallucination in contemporary captioning models, and consider several key questions.
The first question we aim to answer is: \emph{Which models are more prone to hallucination?} We analyze this question on a diverse set of captioning models, spanning different architectures and learning objectives. To measure object hallucination, we propose a new metric, \emph{\metric{} (\metricfullname{})}, which captures image relevance of the generated captions. Specifically, we consider both ground truth object annotations (MSCOCO Object segmentation~\cite{lin2014microsoft}) and ground truth sentence annotations (MSCOCO Captions~\cite{chen2015microsoft}).
Interestingly, we find that models which score best on standard sentence metrics do not always hallucinate less.%

The second question we raise is: \emph{What are the likely causes of hallucination?} %
While hallucination may occur due to a number of reasons, we believe the top factors include visual misclassification and over-reliance on language priors. The latter may result in memorizing which words ``go together'' regardless of image content, which may lead to poor generalization, once the test distribution is changed. We propose \emph{image and language model consistency} scores to investigate this issue, %
and find that models which hallucinate more tend to make mistakes consistent with a language model.%

Finally, we ask: \emph{How well do the standard metrics capture hallucination?} It is a common practice to rely on automatic sentence metrics, e.g. CIDEr \cite{cider}, to evaluate  captioning performance during development, and few employ human evaluation to measure the final performance of their models. As we largely rely on these metrics, it is important to understand how well they capture the hallucination phenomenon. In Figure \ref{fig:teaser} we show how two sentences, from NBT with hallucination and from TopDown model~\cite{anderson2017bottom} -- without, are scored by the standard metrics. As we see, hallucination is not always appropriately penalized. We find that by using additional ground truth data about the image in the form of object labels, our metric \metric{} allows us to catch discrepancies that the standard captioning metrics cannot fully capture. We then investigate ways to assess object hallucination risk with the standard metrics. Finally, we show that \metric{} is complementary to the standard metrics in terms of capturing human preference.

%% file: method.tex
\section{Caption Hallucination Assessment}

We first introduce our image relevance metric, \emph{CHAIR}, which assesses captions w.r.t. objects that are actually in an image. It is used as a main tool in our evaluation. Next we discuss the notions of \emph{image and language model consistency}, which we use to reason about the causes of hallucination. %

\subsection{The \metric{} Metric}
\label{sec:metric}
To measure object hallucination, we propose the \emph{\metric{} (\metricfullname{})} metric, which calculates what proportion of words generated are actually in the image according to the ground truth sentences and object segmentations. This metric has two variants: per-instance, or what fraction of object instances are hallucinated (denoted as \metric{}i), and per-sentence, or what fraction of sentences include a hallucinated object (denoted as \metric{}s):%
$$\text{CHAIR}_{i} = \frac{ \vert\{ \text{hallucinated objects}\}\vert }{\vert\{\text{all objects mentioned}\}\vert }$$
$$\text{CHAIR}_{s} = \frac{ \vert\{ \text{sentences with hallucinated object}\}\vert }{\vert\{\text{  all sentences}\}\vert }$$

For easier analysis, we restrict our study to the 80 MSCOCO objects which appear in the MSCOCO segmentation challenge.
To determine whether a generated sentence contains hallucinated objects, we first tokenize each sentence and then singularize each word. 
We then use a list of synonyms for MSCOCO objects (based on the list from~\citet{lu2018neural}) to map words (e.g., ``player'') to MSCOCO objects (e.g., ``person''). 
Additionally, for sentences which include two word compounds (e.g., ``hot dog'') we take care that other MSCOCO objects (in this case ``dog'') are not incorrectly assigned to the list of MSCOCO objects in the sentence.
For each ground truth sentence, we determine a list of MSCOCO objects in the same way.
The MSCOCO segmentation annotations are used by simply relying on the provided object labels.

We find that considering both sources of annotation is important.
For example, MSCOCO contains an object ``dining table'' annotated with segmentation maps. However, humans refer to many different kinds of objects as ``table'' (e.g., ``coffee table'' or ``side table''), though these objects are not annotated as they are not specifically ``dining table''.
By using sentence annotations to scrape ground truth objects, we account for variation in how human annotators refer to different objects.
Inversely, we find that frequently humans will not mention all objects in a scene. 
Qualitatively, we observe that both annotations are important to capture hallucination.
Empirically, we verify that using only segmentation labels or only reference captions leads to higher hallucination (and practically incorrect) rates.

\subsection{Image Consistency}
\label{sec:img_consistency}
We define a notion of \textit{image consistency}, or how consistent errors from the captioning model are with a model which predicts objects based on an image alone.
To measure image consistency for a particular generated word, we train an image model and record $P(w|I)$ or the probability of predicting the word given only the image.
To score the image consistency of a caption we use the average of $P(w|I)$ for all MSCOCO objects, where
higher values mean that errors are \textit{more} consistent with the image model.
Our image model is a multi-label classification model with labels corresponding to MSCOCO objects (labels determined the same way as is done for \metric{}) which shares the visual features with the caption models.

\subsection{Language Consistency}
\label{sec:lang_consistency}
We also introduce a notion of \textit{language consistency}, i.e. how consistent errors from the captioning model are with a model which predicts words based only on previously generated words. We train an LSTM~\cite{hochreiter1997long} based language model which predicts a word $w_t$ given previous words $w_{0:t-1}$ 
 on MSCOCO data. 
We report language consistency as $1/R(w_t)$ where $R(w_t)$ is the rank of the predicted word in the language model.
Again, for a caption we report average rank across all MSCOCO objects in the sentence and higher language consistency implies that errors are \textit{more} consistent with the language model.

\begin{figure}[t]
\centering
\includegraphics[width=\linewidth]{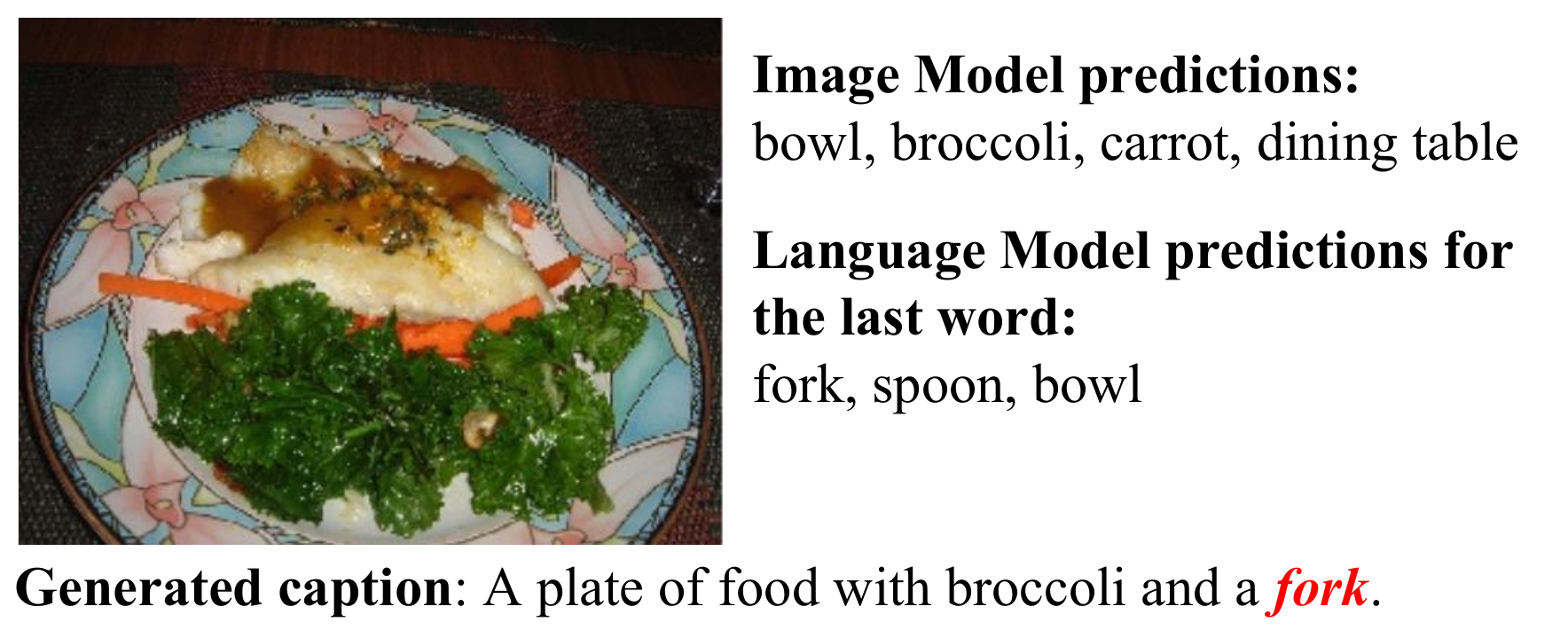}
\caption{\label{fig:lm_im} Example of image and language consistency. The hallucination error (``fork'') is more consistent with the Language Model.}
\end{figure}

We illustrate image and language consistency in Figure~\ref{fig:lm_im}, i.e. the hallucination error (``fork'') is more consistent with the Language Model predictions than with the Image Model predictions. We use these consistency measures in Section \ref{sec:causes} to help us investigate the causes of hallucination.

%% file: experiments.tex
\section{Evaluation}
In this section we present the findings of our study, where we aim to answer the questions posed in Section~\ref{sec:intro}: \emph{Which models are more prone to hallucination? 
What are the likely causes of hallucination? How well do the standard metrics capture hallucination?}

\subsection{Baseline Captioning Models}
We compare object hallucination across a wide range of models.
We define two axes for comparison: model architecture and learning objective. 

\textit{Model architecture.}
Regarding model architecture, we consider models both with and without attention mechanisms.
In this work, we use ``attention'' to refer to any mechanism which learns to focus on different image regions, whether image regions be determined by a high level feature map, or by object proposals from a trained detector.
All models are end-to-end trainable and use a recurrent neural network (LSTM~\cite{hochreiter1997long} in our case) to output text.
For non-attention based methods we consider the \textbf{FC model} from~\citet{rennie2017cvpr} which incorporates visual information by initializing the LSTM hidden state with high level image features. 
We also consider \textbf{LRCN}~\cite{donahue2015long} which considers visual information at each time step, as opposed to just initializing the LSTM hidden state with extracted features.

For attention based models, we consider  \textbf{Att2In}~\cite{rennie2017cvpr}, which is similar to the original attention based model proposed by~\cite{xu2015show}, except the image feature is only input into the cell gate as this was shown to lead to better performance.
We then consider the attention model proposed by~\cite{anderson2017bottom} which proposes a specific ``top-down attention'' LSTM as well as a ``language'' LSTM.  %
Generally attention mechanisms operate over high level convolutional layers.
The attention mechanism from~\cite{anderson2017bottom} can be used on such feature maps, but \citeauthor{anderson2017bottom} also consider feature maps corresponding to object proposals from a detection model.
We consider both models, denoted as \textbf{TopDown} (feature map extracted from high level convolutional layer) and \textbf{TopDown-BB} (feature map extracted from object proposals from a detection model).
Finally, we consider the recently proposed \textbf{Neural Baby Talk (NBT)} model~\cite{lu2018neural} which explicitly uses object detections (as opposed to just bounding boxes) for sentence generation.%

\begin{table*}
\centering
\small
\begin{tabular}{ l | c|r r r r r|r r r r r} 
 \toprule
 &  & \multicolumn{5}{c|}{Cross Entropy} &  \multicolumn{5}{c}{Self Critical} \\
  Model & Att. & S & M & C & CHs & CHi & S & M & C & CHs & CHi\\ 
  \hline
 LRCN* & & 17.0	& 23.9	& 90.8	& 17.7	& 12.6 & 16.9	& 23.5	& 93.0	& 17.7	& 12.9 \\
 FC* & & 17.9	&24.9	&95.8	& 15.4	& 11.0  & 18.4	&25.0	&103.9	&	14.4 & 10.1\\
 Att2In* & \checkmark & 18.9&	25.8	& 102.0& 10.8	&	7.9 & 19.0	&25.7	&106.7	&	12.2 & 8.4\\
 TopDown*& \checkmark & 19.9	&26.7	&107.6	&	8.4 & 6.1   & 20.4	&27.0	&117.2	& 13.6	& 8.8 \\
 \midrule
 TopDown-BB $^\dagger$ & \checkmark & 20.4 &	27.1	& 113.7	& 8.3	& 5.9 &  21.4	&27.7	&120.6	&	10.4 & 6.9  \\
 NBT $^\dagger$ & \checkmark & 19.4 & 26.2 & 105.1 &  7.4 & 5.4  & - & - & - & - \\
 \midrule 
  &  & \multicolumn{5}{c|}{Cross Entropy} &  \multicolumn{5}{c}{GAN} \\
 GAN $^\ddagger$ & & 18.7 &	25.7	& 100.4	& 10.7	&  7.7 & 16.6 &	22.7 &	79.3 &	8.2 & 6.5	 \\
 \bottomrule
\end{tabular}
\caption{\small Hallucination analysis on the Karpathy Test set: Spice (S), CIDEr (C) and METEOR (M) scores across different image captioning models as well as \metric{}s (sentence level, CHs) and \metric{}i (instance level, CHi).
All models are generated with beam search (beam size=5).  * are trained/evaluated within the same implementation~\cite{luo2018discriminability}, $^\dagger$ are trained/evaluated with implementation publicly released with corresponding papers, and $^\ddagger$ sentences obtained directly from the author. For discussion see Section~\ref{sec:sota_hallucination}.}
\label{tab:hallucination}
\end{table*}

\textit{Learning objective.}
All of the above models are trained with the standard \emph{cross entropy} (CE) loss as well as the \emph{self-critical} (SC) loss proposed by~\citet{rennie2017cvpr} (with an exception of NBT, where only the CE version is included).
The SC loss directly optimizes the CIDEr metric with a reinforcement learning technique.
We additionally consider a model trained with a \emph{GAN} loss~\cite{shetty2017speaking} (denoted \textbf{GAN}), which applies adversarial training to obtain more diverse and ``human-like'' captions, and their respective non-GAN baseline with the CE loss.

\textit{TopDown deconstruction.} %
To better evaluate how each component of a model might influence hallucination, we ``deconstruct'' the TopDown model by gradually removing components until it is equivalent to the FC model. 
The intermediate networks are \textit{NoAttention}, in which the attention mechanism is replaced by mean pooling, \textit{NoConv} in which spatial feature maps are not input into the network (the model is provided with fully connected feature maps), \textit{SingleLayer} in which only one LSTM is included in the model, and finally, instead of inputting visual features at each time step, visual features are used to initialize the LSTM embedding as is done in the FC model.  By deconstructing the TopDown model in this way, we ensure that model design choices and hyperparameters do not confound results.

\textit{Implementation details.}
All the baseline models employ features extracted from the fourth layer of ResNet-101~\cite{he2016deep}, except for the GAN model which employs ResNet-152.
Models without attention traditionally use fully connected layers as opposed to convolutional layers.
However, as ResNet-101 does not have intermediate fully connected layers, it is standard to average pool convolutional activations and input these features into non-attention based description models.
Note that this means the difference between the \textit{NoAttention} and \textit{NoConv} model is that the \textit{NoAttention} model learns a visual embedding of spatial feature maps as opposed to relying on pre-pooled feature maps.
All models except for TopDown-BB, NBT, and GAN are implemented in the same open source framework from~\citet{luo2018discriminability}.\footnote{\url{https://github.com/ruotianluo/self-critical.pytorch}}

\textit{Training/Test splits.}
We evaluate the captioning models on two MSCOCO splits. First, we consider the split from Karpathy et al.~\cite{karpathy2015deep}, specifically in that case the models are trained on the respective Karpathy Training set, tuned on Karpathy Validation set and the reported numbers are on the Karpathy Test set.
We also consider the \emph{Robust} split, introduced in \cite{lu2018neural}, which provides a compositional split for MSCOCO. Specifically, it is ensured that the object pairs present in the training, validation and test captions do not overlap. %
In this case the captioning models are trained on the Robust Training set, tuned on the Robust Validation set and the reported numbers are on the Robust Test set.
\subsection{Which Models Are More Prone To Hallucination?}
\label{sec:sota_hallucination}

We first present how well competitive models perform on our proposed \metric{} metric (Table~\ref{tab:hallucination}). We report \metric{}  at sentence-level and at instance-level (CHs and CHi in the table).
In general, we see that models which perform better on standard evaluation metrics, perform better on \metric{}, though this is not always true.
In particular, models which optimize for CIDEr frequently hallucinate more.
Out of all generated captions on the Karpathy Test set, anywhere between 7.4\% and 17.7\% include a hallucinated object.
When shifting to more difficult training scenarios in which new combinations of objects are seen at test time, hallucination consistently increases (Table~\ref{tab:hallucination-robust}).

\textit{Karpathy Test set.}  Table~\ref{tab:hallucination} presents object hallucination 
on the Karpathy Test set.
All sentences are generated using beam search and a beam size of 5.
We note a few important trends. First, models with attention tend to perform better on the \metric{} metric than models without attention.
As we explore later, this is likely because they have a better understanding of the image.
In particular, methods that incorporate bounding box attention (as opposed to relying on coarse feature maps), consistently have lower hallucination as measured by our \metric{} metric.
Note that the NBT model does not perform as well on standard captioning metrics as the TopDown-BB model but has lower hallucination.
This is perhaps because bounding box proposals come from the MSCOCO detection task and are thus ``in-domain'' as opposed to the TopDown-BB model which relies on proposals learned from the Visual Genome~\cite{krishna2017visual} dataset.
Second, frequently training models with the self-critical loss actually increases the amount of hallucination. One hypothesis is that CIDEr does not penalize object hallucination sufficiently, leading to both increased CIDEr and increased hallucination.
Finally, the LRCN model has a higher hallucination rate than the FC model, indicating that inputting the visual features only at the first step, instead of at every step, leads to more image relevant captions.

We also consider a GAN based model ~\cite{shetty2017speaking} in our analysis. We include a baseline model (trained with CE) as well as a model trained with the GAN loss.\footnote{Sentences were procured directly from the authors.}
Unlike other models, the GAN model uses a stronger visual network (ResNet-152) which could explain the lower hallucination rate for both the baseline and the GAN model.
Interestingly, when comparing the baseline and the GAN model (both trained with ResNet-152), standard metrics decrease substantially, even though human evaluations from ~\cite{shetty2017speaking} demonstrate that sentences are of comparable quality.
On the other hand, hallucination decreases, implying that the GAN loss actually helps decrease hallucination. %
Unlike the self critical loss, the GAN loss encourages sentences to be human-like as opposed to optimizing a metric.
Human-like sentences are not likely to hallucinate objects, and a hallucinated object is likely a strong signal to the discriminator that a sentence is generated, and is not from a human.

\begin{figure*}[t]
\centering
\includegraphics[width=\linewidth]{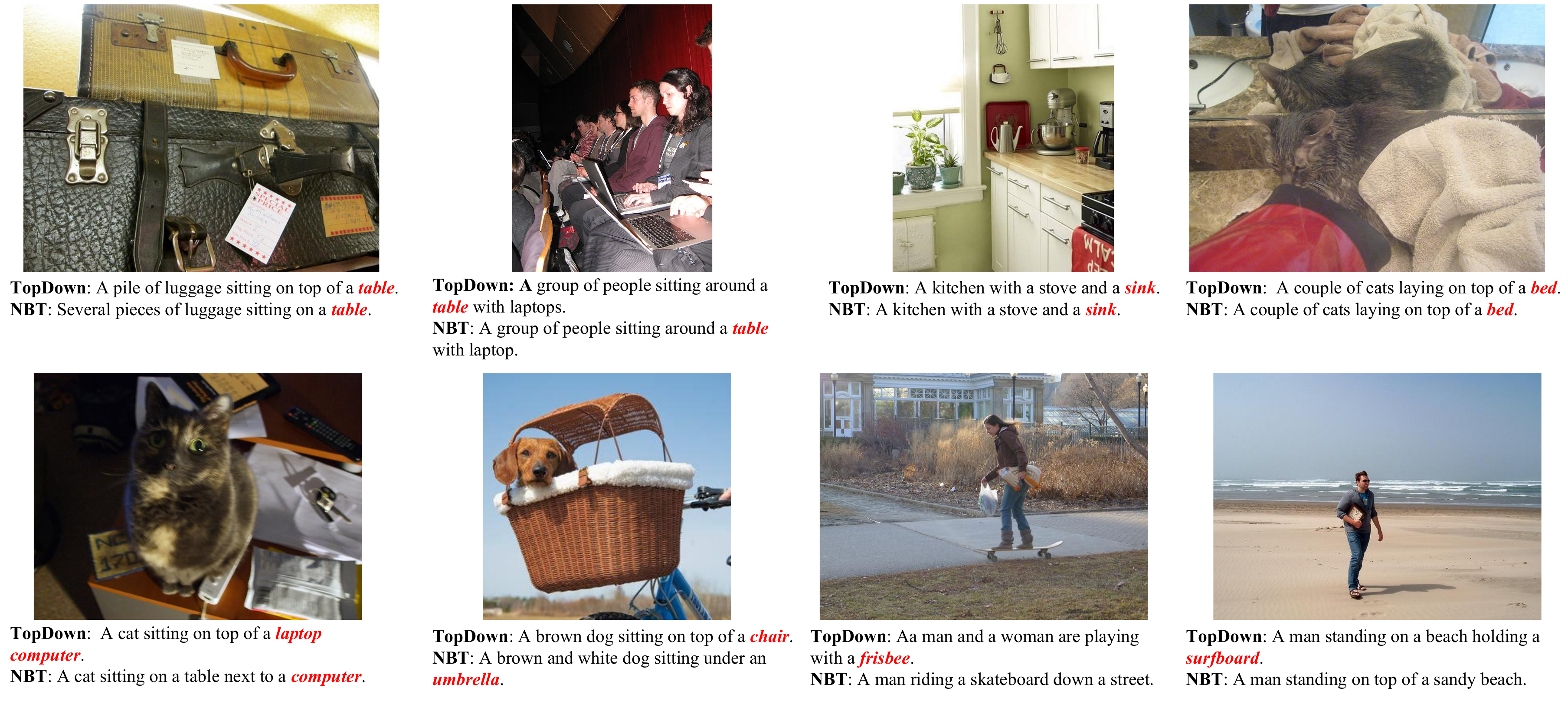}
\caption{\label{fig:qualitative} \small Examples of object hallucination from two state-of-the-art captioning models, TopDown and NBT, see Section~\ref{sec:sota_hallucination}.}
\end{figure*}

We also assess the effect of beam size on \metric{}. We find that generally beam search decreases hallucination. 
We use beam size of 5, and for all models trained with cross entropy, it outperforms lower beam sizes on \metric{}.
However, when training models with the self-critical loss, beam size sometimes leads to worse performance on \metric{}.
For example, on the Att2In model trained with SC loss, a beam size of 5 leads to $12.2$ on \metric{}s and $8.4$ on \metric{}i, while a beam size of 1 leads to $10.8$ on \metric{}s and $8.1$ on \metric{}i.

\begin{table}[t]
\centering
\small
\begin{tabular}{@{}l@{\ }|@{\ }c@{\ }|@{\ }c@{\ \ }c@{\ \ }c@{\ \ }c@{\ \ }c@{\ \ }c@{}}
\toprule
        & Att  & S & M & C & CHs   & CHi     \\
\midrule
FC*      &            & 15.5 & 22.7  & 76.2 & 21.3 & 15.3   \\
Att2In*  & \checkmark  & 16.9 & 24.0   & 85.8 &  14.1 & 10.1    \\
TopDown* & \checkmark  & 17.7 & 24.7  & 89.8 & 11.3 & 7.9    \\
NBT $^\dagger$    & \checkmark   & 18.2  & 24.9   & 93.5  & 6.2 & 4.2    \\
\bottomrule
\end{tabular}
\caption{\small Hallucination Analysis on the Robust Test set: Spice (S), CIDEr (C) and METEOR (M) scores across different image captioning models as well as \metric{}s (sentence level, CHs) and \metric{}i (instance level, CHi). * are trained/evaluated within the same implementation~\cite{luo2018discriminability}, $^\dagger$ are trained/evaluated with implementation publicly released with corresponding papers. All models trained with cross-entropy loss. See Section~\ref{sec:sota_hallucination}.}
\label{tab:hallucination-robust}
\end{table}

\textit{Robust Test set.}
Next we review the hallucination behavior on the Robust Test set (Table~\ref{tab:hallucination-robust}).
For almost all models the hallucination increases on the Robust split (e.g. for TopDown from 8.4\% to 11.3\% of sentences), indicating that the issue of hallucination is more critical in scenarios where test examples can not be assumed to have the same distribution as train examples. We again note that attention is helpful for decreasing hallucination.
We note that the NBT model actually has lower hallucination scores on the robust split.
This is in part because when generating sentences we use the detector outputs provided by~\citet{lu2018neural}.
Separate detectors on the Karpathy test and robust split are not available and the detector has access to images in the robust split during training.
Consequently, the comparison between NBT and other models is not completely fair, but we include the number for completeness.

\begin{figure*}[t]
\centering
\includegraphics[width=\linewidth]{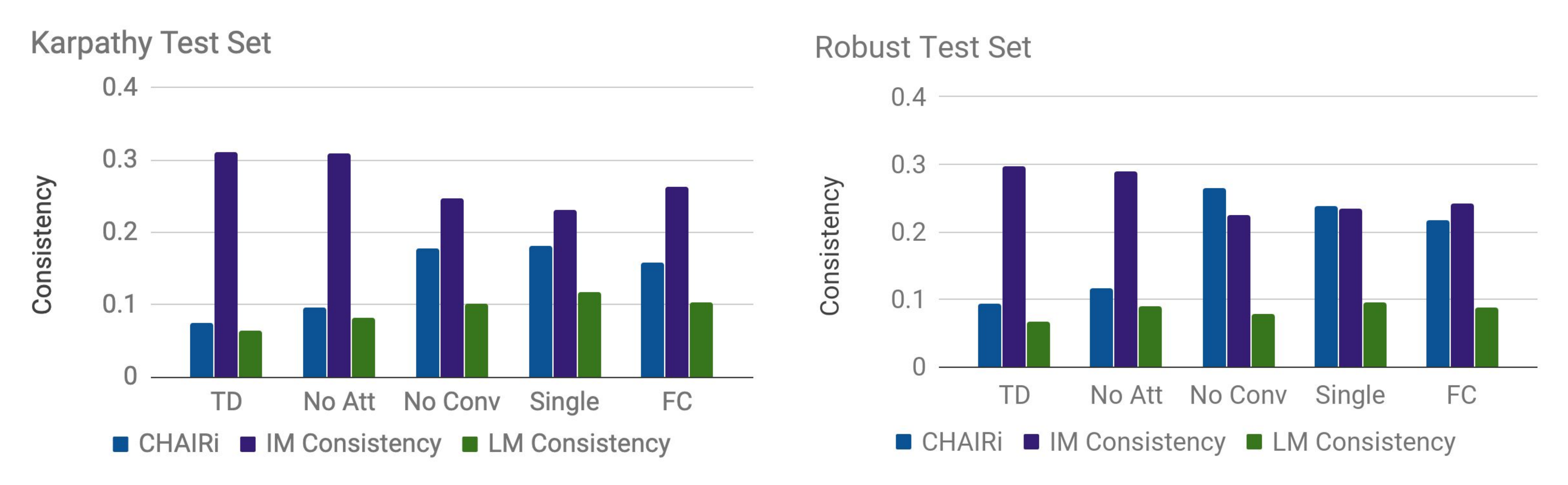}
\caption{\label{fig:consistency} \small Image and Language model consistency (IM, LM) and \metric{}i (instance-level, CHi) on deconstructed TopDown models. Images with less hallucination tend to make errors consistent with the image model, whereas models with more hallucination tend to make errors consistent with the language model, see Section~\ref{sec:causes}.}
\end{figure*}

In addition to the Robust Test set, we also consider a set of MSCOCO in which certain objects are held out, which we call the \emph{Novel Object split}~\cite{hendricks2016deep}.
We train on the training set outlined in~\cite{hendricks2016deep} and test on the Karpathy test split, which includes objects unseen during training.
Similarly to the Robust Test set, we see hallucination increase substantially on this split.
For example, for the TopDown model hallucination increases from 8.4\% to 12.1\% for CHAIRs and 6.0\% to 9.1\% for CHAIRi.%

We find no obvious correlation between the average length of the generated captions and the hallucination rate. Moreover, vocabulary size does not correlate with hallucination either, i.e. models with \emph{more diverse} descriptions may actually \emph{hallucinate less}. We notice that hallucinated objects tend to be mentioned towards \emph{the end of the sentence} (on average at position 6, with average sentence length 9), suggesting that some of the preceding words may have triggered hallucination. We investigate this below.

\paragraph{Which objects are hallucinated and in what context?}
Here we analyze which MSCOCO objects tend to be hallucinated more often and what are the common preceding words and  image context. %
Across all models the super-category \emph{Furniture} is hallucinated most often, accounting for $20 - 50\%$ of all hallucinated objects. Other common super-categories are \emph{Outdoor objects}, \emph{Sports} and \emph{Kitchenware}. On the Robust Test set, \emph{Animals} are often hallucinated. The \emph{dining table} is the most frequently hallucinated object across all models (with an exception of GAN, where \emph{person} is the most hallucinated object). We find that often words like ``sitting'' and ``top'' precede the ``dining table'' hallucination, implying the two common scenarios: a person ``sitting at the table'' and an object ``sitting on top of the table'' (Figure~\ref{fig:qualitative}, row 1, examples 1, 2). Similar observations can be made for other objects, e.g. word ``kitchen'' often precedes ``sink'' hallucination (Figure~\ref{fig:qualitative}, row 1, example 3) and ``laying'' precedes ``bed'' (Figure~\ref{fig:qualitative}, row 1, example 4). 
At the same time, if we look at which objects are actually present in the image (based on MSCOCO object annotations), we can similarly identify that presence of a ``cat'' co-occurs with hallucinating a ``laptop'' (Figure~\ref{fig:qualitative}, row 2, example 1), a ``dog'' -- with a ``chair'' (Figure~\ref{fig:qualitative}, row 2, example 2) etc. In most cases we observe that the hallucinated objects appear in the relevant scenes (e.g. ``surfboard'' on a beach), but there are cases where objects are hallucinated out of context (e.g. ``bed'' in the bathroom, Figure~\ref{fig:qualitative}, row 1, example 4).

\subsection{What Are The Likely Causes Of Hallucination?}
\label{sec:causes}

In this section we investigate the likely causes of object hallucination. We have earlier described how we deconstruct the TopDown model to enable a controlled experimental setup. We rely on the deconstructed TopDown models to analyze the impact of model components on hallucination.

First, we summarize the hallucination analysis on the deconstructed TopDown models (Table~\ref{tab:deconstruction}).
Interestingly, the \textit{NoAttention} model does not do substantially worse than the full model (w.r.t. sentence metrics and \metric{}).
However, removing Conv input (\textit{NoConv} model) and relying only on FC features, decreases the performance dramatically.
This suggests that much of the gain in attention based models is primarily due to \emph{access to feature maps with spatial locality}, not the actual attention mechanism.
Also, similar to LRCN vs. FC in Table~\ref{tab:hallucination}, initializing the LSTM hidden state with image features, as opposed to inputting image features at each time step, leads to lower hallucination (\textit{Single Layer} vs. \textit{FC}). 
This is somewhat surprising, as a model which has access to image information at each time step should be less likely to ``forget'' image content and hallucinate objects. However, it is possible that models which include image inputs at each time step with no access to spatial features overfit to the visual features.%

\begin{table}
\centering
\resizebox{\linewidth}{!}{
\begin{tabular}{l|rrrrr}
\toprule
Karpathy Split & S & M & C  & CHs & CHi      \\
\midrule
TD           & 19.5 & 26.1 & 103.4 & 10.8        &  7.5  \\
No Attention & 18.8 & 25.6 & 99.7 & 14.2        &  9.5  \\
No Conv      & 15.7 & 22.9 & 81.3 & 25.7        &  17.8  \\
Single Layer & 15.5 & 22.7 & 80.2 & 25.7        &  18.2 \\
FC           & 16.4 & 23.3 & 85.1  & 23.6        &  15.8  \\
\end{tabular}
}
\caption{\small Hallucination analysis on deconstructed TopDown models with sentence metrics SPICE (S), METEOR (M), and CIDEr (C), \metric{}s (sentence level, CHs) and \metric{}i (instance level, CHi). See Section~\ref{sec:causes}.}
\label{tab:deconstruction}
\end{table}

Now we investigate what causes hallucination using the deconstructed TopDown models and the  \textit{image consistency} and \textit{language consistency} scores, introduced in Sections~\ref{sec:img_consistency} and \ref{sec:lang_consistency} which capture how consistent the hallucinations errors are with image- / language-only models.

Figure~\ref{fig:consistency} shows the \metric{} metric, image consistency and language consistency for the deconstructed TopDown models on the Karpathy Test set (left) and the Robust Test set (right).
We note that models with \textit{less} hallucination tend to make errors consistent with the image model, whereas models with \textit{more} hallucination tend to make errors consistent with the language model. 
This implies that models with less hallucination are better at integrating knowledge from an image into the sentence generation process.
When looking at the Robust Test set, Figure~\ref{fig:consistency} (right), which is more challenging, as we have shown earlier, 
we see that image consistency \emph{decreases} when comparing to the same models on the Karpathy split, whereas language consistency is similar across all models trained on the Robust split. 
This is perhaps because the Robust split contains novel compositions of objects at test time, and all of the models are heavily biased by language.  %

Finally, we measure image and language consistency during training for the FC model and note that at the beginning of training errors are more consistent with the language model, whereas towards the end of training, errors are more consistent with the image model. %
This suggests that models first learn to produce fluent language before learning to incorporate visual information. %

\begin{figure}[t]
\centering
\includegraphics[width=\linewidth]{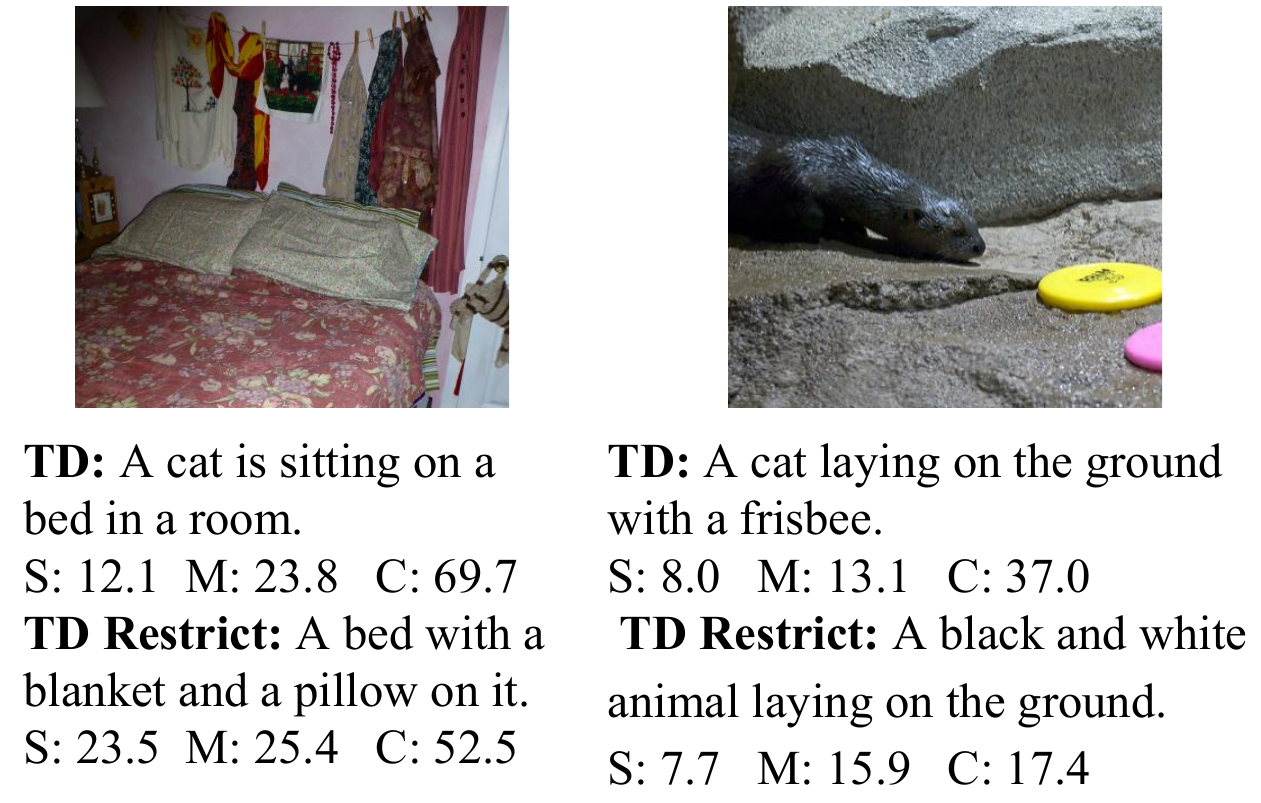}
\caption{\label{fig:restrict} \small Examples of how TopDown (TD) sentences change when we enforce that objects cannot be hallucinated: SPICE (S), Meteor (M), CIDEr (C), see Section~\ref{sec:metrics_analysis}.}
\vspace{-0.5cm}
\end{figure}

\subsection{How Well Do The Standard Metrics Capture Hallucination?}
\label{sec:metrics_analysis}

In this section we analyze how well SPICE~\cite{spice}, METEOR~\cite{meteor}, and CIDEr~\cite{cider} capture hallucination. All three metrics do penalize sentences for mentioning incorrect words, either via an F score (METEOR and SPICE) or cosine distance (CIDEr). However, if a caption mentions enough words correctly, it can have a high METEOR, SPICE, or CIDEr score while still hallucinating specific objects.

Our first analysis tool is the TD-Restrict model. This is a  modification of the TopDown model, where we enforce that MSCOCO objects which are not present in an image are \emph{not generated} in the caption. We determine which words refer to objects absent in an image following our approach in Section~\ref{sec:metric}. We then set the log probability for such words to a very low value. We generate sentences with the TopDown and TD-Restrict model with beam search of size $1$, meaning all words produced by both models are the same, until the TopDown model produces a hallucinated word.

We compare which scores are assigned to such captions in Figure~\ref{fig:restrict}. TD-Restrict generates captions that do not contain hallucinated objects, while TD hallucinates a ``cat'' in both cases. In Figure~\ref{fig:restrict} (left) we see that CIDEr scores the more correct caption much lower. %
In Figure~\ref{fig:restrict} (right), the TopDown model incorrectly calls the animal a ``cat.'' Interestingly, it then correctly identifies the ``frisbee,'' which the TD-Restrict model fails to mention, leading to lower SPICE and CIDEr.

\begin{table}[t]
\centering
\small
\begin{tabular}{l|c|cccc}
\toprule
        &  CIDEr & METEOR & SPICE    \\
\midrule
FC      & 0.258 & 0.240  & 0.318 \\
Att2In  & 0.228 &  0.210  & 0.284 \\
TopDown & 0.185  & 0.168  & 0.215 \\
\bottomrule
\end{tabular}
\caption{\small Pearson correlation coefficients between 1-CHs and CIDEr, METEOR, and SPICE scores, see Section~\ref{sec:metrics_analysis}.}
\label{tab:metric-correlation}
\end{table}

In Table~\ref{tab:metric-correlation} we compute Pearson correlation coefficient between individual sentence scores and the \emph{absence} of hallucination, i.e. $1-$\metric{}s; %
we find that SPICE consistently correlates higher with $1-$\metric{}s. E.g., for the FC model the correlation for SPICE is 0.32, while for METEOR and CIDEr -- around 0.25.%

\begin{figure}[t]
\centering
\includegraphics[width=0.95\linewidth]{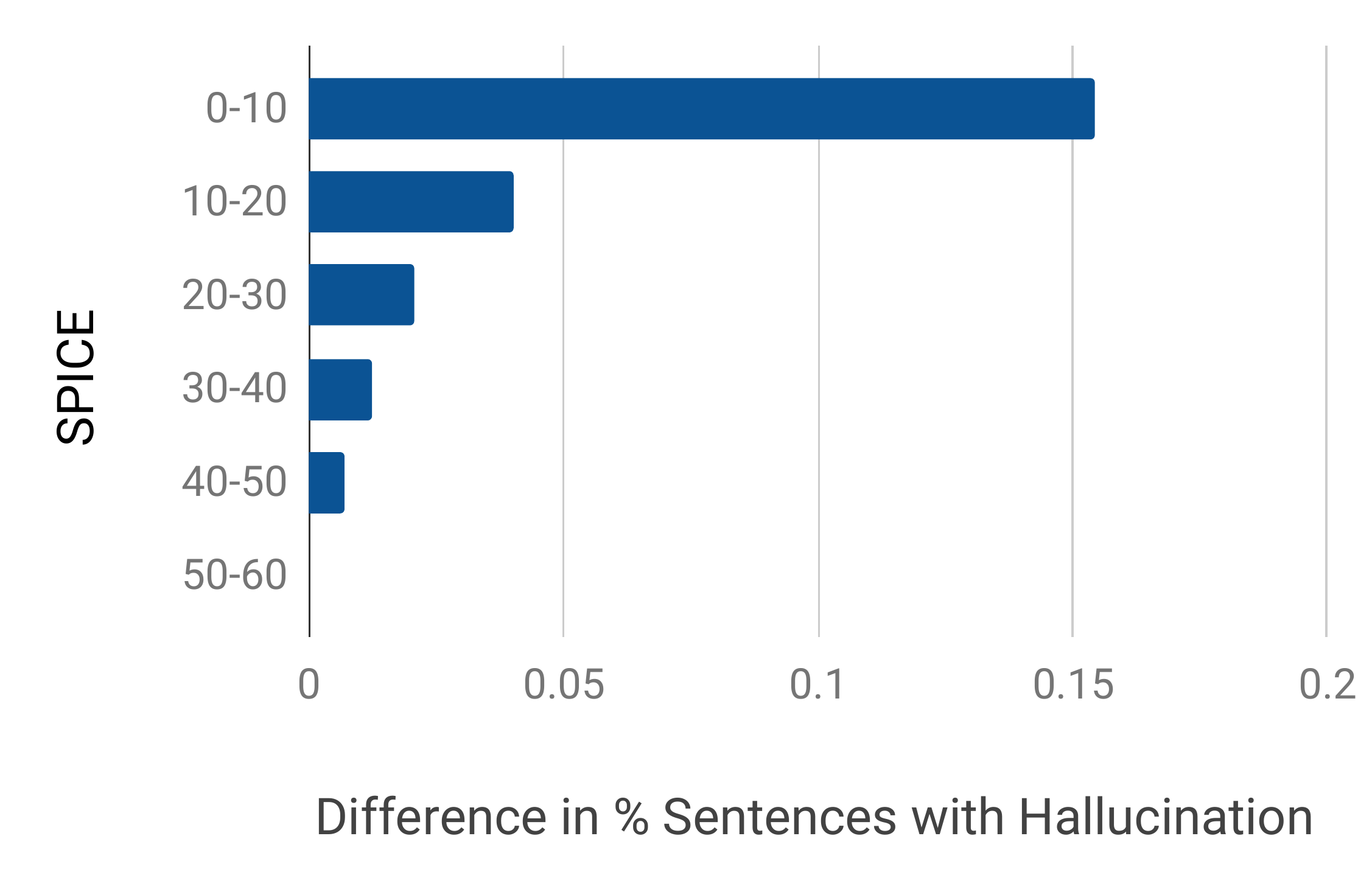}
\caption{\label{fig:metrics} \small Difference in percentage of sentences with \emph{no} hallucination for TopDown and FC models when SPICE scores fall into specific ranges. For sentences with low SPICE scores, the hallucination is generally larger for the FC model, even though the SPICE scores are similar, see Section~\ref{sec:metrics_analysis}.}
\end{figure}

We further analyze the metrics in terms of their predictiveness of hallucination risk. Predictiveness means that a certain score should imply a certain percentage of hallucination. Here we show the results for SPICE and the captioning models FC and TopDown. For each model and a score interval (e.g. $10-20$) we compute the percentage of captions \emph{without} hallucination ($1-$\metric{}s). We  plot the difference between the percentages from both models (TopDown - FC) in Figure~\ref{fig:metrics}. 
Comparing the models, we note that even when scores are similar (e.g., all sentences with SPICE score in the range of $10-20$), the TopDown model has fewer sentences with hallucinated objects.
We see similar trends across other metrics.
Consequently, object hallucination can \textit{not} be always predicted based on the traditional sentence metrics.

\paragraph{Is CHAIR complementary to standard metrics?}
In order to measure usefulness of our proposed metrics, we have conducted the following human evaluation (via the Amazon Mechanical Turk). We have randomly selected 500 test images and respective captions from 5 models: non-GAN baseline, GAN, NBT, TopDown and TopDown - Self Critical. The AMT workers were asked to score the presented captions w.r.t. the given image based on their preference. %
They could score each caption from 5 (very good) to 1 (very bad). We did not use ranking, i.e. different captions could get the same score; each image was scored by three annotators, and the average score is used as the final human score. For each image we consider the 5 captions from all models and their corresponding sentence scores (METEOR, CIDEr, SPICE). We then compute Pearson correlation between the human scores and sentence scores; we also consider a simple combination of sentence metrics and 1-CHAIRs or 1-CHAIRi by summation. The final correlation is computed by averaging across all 500 images. The results are presented in Table~\ref{tab:correlation_human}. Our findings indicate that a simple combination of CHAIRs or CHAIRi with the sentence metrics leads to an increased correlation with the human scores, showing the usefulness and complementarity of our proposed metrics.

\begin{table}[t]
\centering
\small
\resizebox{\linewidth}{!}{
\begin{tabular}{l|ccccc}
\toprule
&  Metric &  Metric & Metric\\
&   &  +(1-CHs) & +(1-CHi)\\
\midrule
METEOR & 0.269 & 0.299 & 0.304 \\
CIDEr  & 0.282 & 0.321 & 0.322 \\
SPICE  & 0.248 & 0.277 & 0.281 \\
\end{tabular}
}
\caption{\small Pearson correlation coefficients between individual/combined metrics and human scores. See Section~\ref{sec:metrics_analysis}.}
\label{tab:correlation_human}
\end{table}

\paragraph{Does hallucination impact generation of other words?} 
Hallucinating objects impacts sentence quality not only because an object is predicted incorrectly, but also because the hallucinated word impacts generation of other words in the sentence. Comparing the sentences generated by TopDown and TD-Restrict allows us to analyze this phenomenon. We find that after the hallucinated word is generated, the following words in the sentence are different 47.3\% of the time. This implies that hallucination impacts sentence quality beyond simply naming an incorrect object.
We observe that one hallucination may lead to another, e.g. hallucinating a ``cat'' leading to hallucinating a ``chair'', hallucinating a ``dog'' -- to a ``frisbee''.

%% file: discussion.tex
\section{Discussion}

In this work we closely analyze hallucination in object captioning models.
Our work is similar to other works which attempt to characterize flaws of different evaluation metrics~\cite{kilickaya2016re}, though we focus specifically on hallucination.
Likewise, our work is related to other work which aims to build better evaluation tools (\cite{cider}, ~\cite{spice}, ~\cite{cui2018learning}).
However, we focus on carefully quantifying and characterizing one important type of error: object hallucination.

A significant number of objects are hallucinated in current captioning models (between 5.5\% and 13.1\% of MSCOCO objects).
Furthermore, hallucination does not always agree with the output of standard captioning metrics. 
For instance, the popular self critical loss increases CIDEr score, but also the amount of hallucination.
Additionally, we find that given two sentences with similar CIDEr, SPICE, or METEOR scores from two different models, the number of hallucinated objects might be quite different.
This is especially apparent when standard metrics assign a low score to a generated sentence.
Thus, for challenging caption tasks on which standard metrics are currently poor (e.g., the LSMDC dataset~\cite{rohrbach2017movie}), the \metric{} metric might be helpful to tease apart the most favorable model. Our results indicate that \metric{} complements the standard sentence metrics in capturing human preference.

Additionally, attention lowers hallucination, but it appears that much of the gain from attention models is due to access to the underlying convolutional features as opposed the attention mechanism itself.
Furthermore, we see that models with stronger \emph{image consistency} frequently hallucinate fewer objects, suggesting that strong visual processing is important for avoiding hallucination.

Based on our results, we argue that the design and training of captioning models should be guided not only by cross-entropy loss or standard sentence metrics, but also by image relevance.
Our \metric{} metric gives a way to evaluate the phenomenon of hallucination, but other image relevance metrics e.g. those that incorporate missed salient objects, should also be investigated. We believe that incorporating visual information in the form of ground truth objects in a scene (as opposed to only reference captions) helps us better understand the performance of captioning models.